\documentclass{article}

\usepackage[final]{neurips_2025}

\usepackage[utf8]{inputenc} 
\usepackage[T1]{fontenc}    
\usepackage{hyperref}       
\usepackage{url}            
\usepackage{booktabs}       
\usepackage{amsfonts}       
\usepackage{nicefrac}       
\usepackage{microtype}      
\usepackage{xcolor}         
\usepackage{amsmath,amssymb,amsfonts}
\usepackage{graphicx}
\usepackage{textcomp}
\usepackage{xcolor}
\def\BibTeX{{\rm B\kern-.05em{\sc i\kern-.025em b}\kern-.08em
    T\kern-.1667em\lower.7ex\hbox{E}\kern-.125emX}}

\usepackage{booktabs} 
\usepackage{import}
\usepackage{graphicx}
\usepackage{epstopdf}
\usepackage{amsmath}
\usepackage{subfigure}
\usepackage{diagbox}
\usepackage{float}
\graphicspath{{Fig/}}
\usepackage{booktabs} 
\usepackage{algorithm}
\usepackage{algorithmicx}
\usepackage{algpseudocode}
\usepackage{paralist}
\usepackage{multirow}
\usepackage{graphicx}
\usepackage{subfigure}
\usepackage{color}
\usepackage{hhline}
\usepackage{xcolor}
\usepackage{color,soul}
\usepackage{array}
\usepackage{diagbox}
\usepackage{hyperref}
\usepackage{verbatim} 
\usepackage{colortbl}
\usepackage{upgreek}

\usepackage{enumitem}
\usepackage{gensymb} 
\usepackage{ulem} 
\usepackage{amsmath,bm}
\usepackage{amssymb}
\usepackage{pifont}
\usepackage{filecontents}
\algdef{SE}[DOWHILE]{Do}{doWhile}{\algorithmicdo}[1]{\algorithmicwhile\ #1}%
\usepackage{amsmath, amssymb, amsthm}
\usepackage{algorithm}

\definecolor{deep_pink}{HTML}{D97692}

\title{Closer to Language than Steam: \\ AI as the Cognitive Engine of a New Productivity Revolution}

\author{%
  Xinmin Fang  \\
  Department of Computer Science and Engineering\\
  University of Colorado Denver (CU Denver)\\
  Denver, CO 80204 \\
  \texttt{xinmin.fang@ucdenver.edu} \\
  \And
  Lingfeng Tao \\
  Department of Robotics and Mechatronics \\
  Kennesaw State University (KSU)\\
  Atlanta, GA, 30144 \\
  \texttt{ltao2@kennesaw.edu} \\
  \AND
  Zhengxiong Li \thanks{Disclaimer: This working paper is intended solely for academic discussion and exploratory analysis. It does not constitute investment advice or serve as a basis for any financial decision-making. All data and examples referenced are derived from publicly available media sources and industry reports. \\
  * This work is partially supported by US NSF Awards \#2426469 and \#2426470.
  } \\
  Department of Computer Science and Engineering \\
  University of Colorado Denver (CU Denver)\\
  Denver, CO 80204 \\
  \texttt{zhengxiong.li@ucdenver.edu} \\
}

\begin{document}

\maketitle
\begin{abstract}
Artificial Intelligence (AI) is reframed as a cognitive engine driving a novel productivity revolution distinct from the Industrial Revolution's physical thrust. This paper develops a theoretical framing of AI as a cognitive revolution akin to written language - a transformative augmentation of human intellect rather than another mechanized tool. We compare AI's emergence to historical leaps in information technology to show how it amplifies knowledge work. Examples from various domains demonstrate AI's impact as a driver of productivity in cognitive tasks. We adopt a multidisciplinary perspective combining computer science advances with economic insights and sociological perspectives on how AI reshapes work and society. Through conceptual frameworks, we visualize the shift from manual to cognitive productivity. \textbf{Our central argument is that AI functions as an engine of cognition - comparable to how human language revolutionized knowledge - heralding a new productivity paradigm.} We discuss how this revolution demands rethinking of skills, organizations, and policies. This paper, balancing academic rigor with clarity, concludes that AI's promise lies in complementing human cognitive abilities, marking a new chapter in productivity evolution.
\end{abstract}

\section{Introduction}
Over two centuries ago, the steam engine catalyzed the Industrial Revolution by augmenting human physical power, mechanizing labor, and driving unprecedented productivity gains in manufacturing and transportation. Today, we stand on the brink of an arguably even more profound upheaval - a revolution not of muscles and machines, but of mind and cognition. Artificial Intelligence (AI), especially in its modern incarnations like machine learning and generative AI, is increasingly viewed as a ``steam engine of the mind,'' delivering cognitive superpowers in any task involving information and language
\cite{mckinsey2024genai}
. This new revolution is qualitatively different from the industrial paradigm: instead of automating physical labor, AI automates and amplifies cognitive labor - the domain of thinking, reasoning, and creating. 

Many commentators have drawn parallels between AI's transformative potential and that of past technological breakthroughs. However, an important insight is that AI should be conceptualized less as a better machine and more as a better mind. In other words, AI is to human thought what the steam engine was to human muscle. \textbf{This perspective echoes through history: just as the invention of written language and literacy thousands of years ago extended human memory and communication, AI extends our capacity to process and generate knowledge.} The emergence of writing in ancient civilizations enabled information to be recorded and transmitted across time and space, fundamentally boosting cognitive productivity by allowing humans to build on knowledge cumulatively. Similarly, the printing press in the 15th century democratized information, triggering an ``information revolution'' that vastly expanded literacy, learning, and innovation. AI builds upon and surpasses these earlier cognitive revolutions - it not only stores and disseminates knowledge, but can actively analyze, synthesize, and generate new knowledge on our behalf. As one recent commentary put it, if Gutenberg's press ``released ideas and information from captivity,'' AI now ``accelerates their analysis to produce conclusions that expand the boundaries of knowledge''
\cite{west2018brookings}
. 

\begin{figure}[h]
\centering
\includegraphics[width=0.9\linewidth]{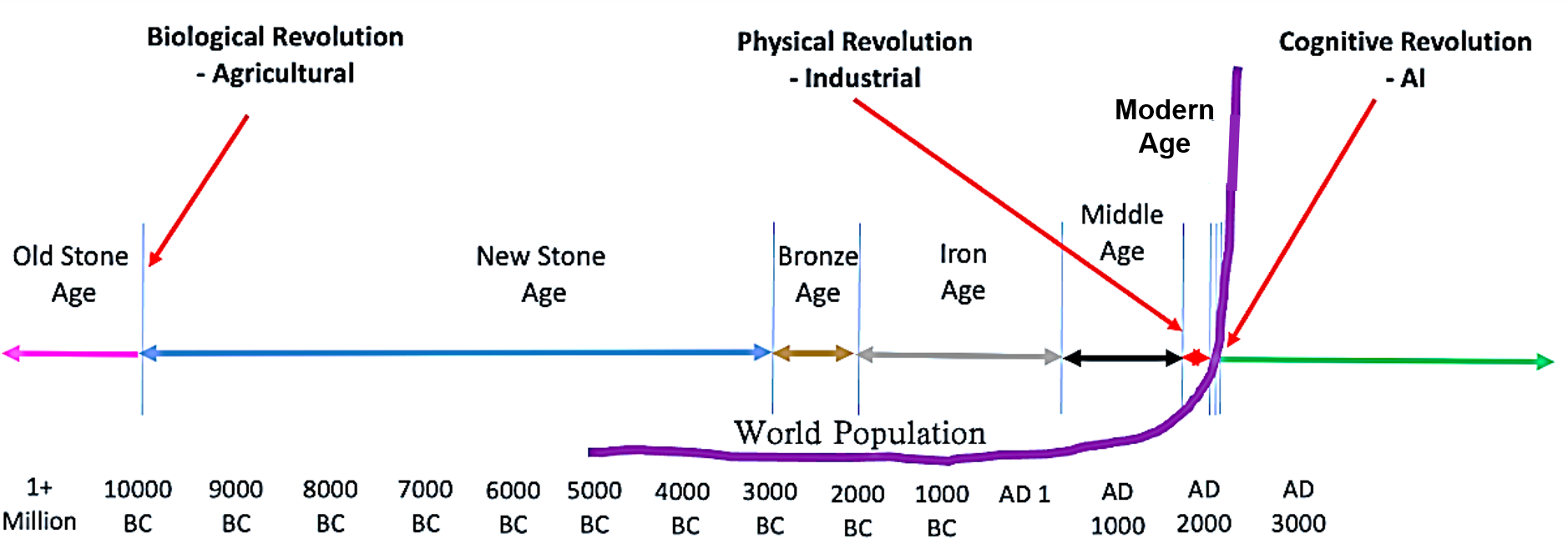}
\caption{The major economic revolutions in human history, highlighting the current AI-driven Cognitive Revolution alongside the earlier Agricultural and Industrial Revolutions. The Agricultural Revolution (around 10,000 BC) was a biological transformation liberating humans from food scarcity through farming; the Industrial Revolution (18th-19th centuries) was a physical transformation liberating humans from onerous manual labor through machinery and steam power. Today's nascent Cognitive Revolution (21st century) is a transformation in the domain of information and knowledge, as AI systems increasingly automate and augment cognitive tasks once done exclusively by humans. This figure illustrates the timeline and fundamental nature of these revolutions, underscoring that AI's impact on cognitive labor may be as pivotal as past revolutions in agriculture and industry. \cite{zhang2023cognitive}}
\label{fig:cog}
\end{figure}

The notion of AI as a cognitive revolution can be framed alongside humanity's previous economic transformations, as shown in Figure~\ref{fig:cog}. Researchers have argued that we are at the start of a third fundamental transformation in human history: following the Agricultural Revolution (which freed humans from food insecurity through farming) and the Industrial Revolution (which freed humans from much physical drudgery through machines), the Cognitive Revolution driven by AI is now liberating humans from routine cognitive labor
\cite{zhang2024preparing}.
In this view, AI's capacity to handle information tasks marks a phase shift as fundamental as those earlier revolutions. Crucially, AI achieves this by energizing all language-based and knowledge-based capabilities in work - including communication, reasoning, analysis, and decision-making
\cite{mckinsey2023productivity}
. Unlike the steam engine's impact on manual labor, AI's impact penetrates any field that relies on information processing and symbolic manipulation (which, in today's economy, is virtually every sector). 

Furthermore, the speed and scale of AI's proliferation exceed those of prior revolutions. Whereas the diffusion of steam power and electricity through economies took decades, modern AI can be distributed globally almost instantaneously via digital networks. Innovations in AI models can reach millions of users in days or weeks thanks to the internet's ubiquity
\cite{mckinsey2024genai}
. For example, OpenAI's ChatGPT - a generative AI conversational agent - acquired over 100 million users just two months after its launch, making it the fastest-adopted consumer application in history
\cite{mehta2023chatgpt}
. This rapid uptake underscores how quickly cognitive tools can spread, heralding swift societal impact. It also signals that the productivity gains from AI could be realized on compressed timescales, if effectively integrated. 

In this paper, we delve deeper into the idea of AI as a cognitive engine powering a new productivity revolution. We begin by elaborating a theoretical framework that compares the AI revolution to the advent of human language and other historic cognitive leaps. We then survey empirical examples across multiple domains - from natural language processing to healthcare and law - where AI is already boosting productivity by taking on cognitive tasks. Next, we adopt an economic lens to discuss how AI fits into growth paradigms, including insights from economists on whether AI will act as a general-purpose technology that augments human labor or simply automates it. We incorporate sociological perspectives on knowledge work and institutional adaptation, examining how workplaces and educational systems must evolve to fully leverage (and wisely manage) AI's capabilities. Throughout, we include visual conceptual frameworks to illustrate the shift from manual to cognitive productivity. Our aim is to blend academic rigor (with references to real studies and data) and narrative clarity to articulate why AI represents not just another tool, but a fundamentally new engine of productivity. 

By reframing AI as akin to human language rather than a steam engine, we highlight that its true power lies in working with symbols, knowledge, and cognitive processes - the very elements that make us human. This perspective carries profound implications. It suggests that the current AI revolution could transform how we work, create value, and organize society on a level comparable to the introduction of writing or the scientific revolution - upheavals in how humans think and produce, not merely how we make things. In the sections that follow, we substantiate this claim and explore its ramifications.


\section{AI as a Cognitive Revolution: Theoretical Framing}
\label{sec:theory}
The defining characteristic of the current AI revolution is its focus on cognitive capacity rather than physical capacity. In theoretical terms, we can liken AI's emergence to a new stage in the evolution of how humans leverage external tools to enhance cognition. Several historical analogies help frame this concept:

1. \textbf{AI and the Invention of Writing}: The emergence of writing systems in ancient Mesopotamia, Egypt, and other early societies was arguably the first great cognitive revolution. Writing allowed humans to offload memory onto durable media (clay tablets, papyrus), to communicate across distance and time, and to build complex administrations and sciences that oral memory alone could not support. This dramatically increased humanity's cognitive productivity - more knowledge could be accumulated, and more complex tasks (like large-scale trade, law, or engineering projects) became possible. Notably, the introduction of writing was initially met with skepticism by some: Plato's dialogues recount Socrates warning that writing might ``annihilate our memory'' by making us reliant on external marks
\cite{anseel2025ai}
. 
Yet, far from diminishing human intellect, literacy vastly amplified it by enabling new forms of thought and culture. AI can be seen as a modern parallel: it is a tool for cognitive offloading and amplification. Just as writing externalized memory, AI externalizes certain reasoning and problem-solving processes. We can delegate routine mental tasks to AI (from recalling facts to solving equations or summarizing texts), thereby freeing human minds to focus on higher-level creativity and strategy. In effect, AI serves as an ``auxiliary brain'' - a cognitive extension of ourselves - which, used properly, could make an individual knowledge worker ``ten times more productive'' than before
\cite{anseel2025ai}
. The difference is that AI goes even further than writing: whereas writing is static storage of human thoughts, AI can actively manipulate and generate new thoughts (e.g. writing code, drafting reports, answering complex questions) in a dynamic, interactive way.

2. \textbf{AI and the Printing Press}: If writing was the first cognitive leap, the printing press was the second. Gutenberg's movable-type press (c. 1440) transformed the productivity of knowledge by replicating information at mass scale. Books that once took months or years to copy by hand could be produced in hours, radically reducing the cost of information dissemination. This had an explosive impact on literacy, scientific exchange, and the Renaissance and Enlightenment that followed - a true information revolution. Today's digital and AI technologies are often compared to Gutenberg in their impact. AI, in particular, not only spreads information but can interpret and synthesize it. A Brookings Institution report phrased it succinctly: ``Artificial intelligence (AI) is the biggest thing since Johannes Gutenberg's 15th century printing press. Gutenberg released ideas and information from captivity; AI accelerates their analysis to produce conclusions that expand the boundaries of knowledge.''
\cite{wheeler2024gutenberg}
. This captures the essence of AI's qualitative difference: it accelerates cognition itself. For instance, an AI system can analyze millions of documents or data points to extract an insight in minutes - something that would take human experts years to accomplish, if it were possible at all. In economic terms, AI dramatically lowers the cost of certain cognitive tasks (like data analysis, translation, optimization) and thereby enables new scales and scopes of intellectual work.

3. \textbf{AI and Language Itself}: The very structure of modern AI - especially large language models like GPT-4 - underscores the idea that AI is fundamentally about language and knowledge manipulation. These models are trained on human language patterns; they operate by predicting and generating linguistically and semantically coherent outputs. In a sense, such AI thinks in human language. This has led some to argue that ``AI is human language, not steam engine'' - highlighting that AI's core competency is working with symbols, abstractions, and meanings, not moving pistons or electrons in the physical world. AI's ``fuel'' is data and information, and its ``output'' is insight or content - much like how human cognitive work is ultimately expressed in language or symbolic form. An AI like GPT can draft an essay, compose music, or hold a conversation - tasks that engage with human language and creativity. It is thus a direct amplifier of human expressive and intellectual capabilities. One researcher described this as ``data is the new oil and AI is the new engine that is transforming data to generate cognitive power'', drawing an analogy between how engines convert fuel to mechanical work and how AI converts data to knowledge work
\cite{zhang2023cognitive}
. The key distinction remains: classical engines amplified muscle power, whereas AI engines amplify mental power.

Synthesizing these perspectives, we define the AI-driven cognitive revolution as a phase in which cognitive tasks - tasks involving perception, understanding, reasoning, and communication - become increasingly automated and enhanced by machines. It is a revolution because it changes the fundamental locus of productivity growth: from expanding human physical capability to expanding human intellectual capability.

Notably, the cognitive revolution driven by AI is unfolding much faster than its historical predecessors and is potentially more ubiquitous. The Agricultural and Industrial Revolutions were initially confined to certain sectors (farming and manufacturing, respectively) and geographic regions before spreading. In contrast, AI is, from the outset, propagating through virtually all industries and across the globe via digital infrastructure. Generative AI, for example, can be applied to customer service, programming, design, education, research - all at once, thanks to its general ability to handle language and knowledge. This generality is why many consider AI a General Purpose Technology (GPT) in the economic sense - akin to the steam engine, electricity, or the computer - but it might better be called a General Purpose Cognitive Technology. Its applications are as broad as human thought itself. 

In the next sections, we provide the mathematical formalization and concrete examples of how this cognitive engine is being put to work in today's world, illustrating the early stages of the productivity revolution it is spawning.

\section{Theoretical Framing: Mathematical Formalization of AI as a Cognitive Revolution}

To formalize the concept of AI as a cognitive productivity revolution, we develop mathematical models that capture the shifts in cognitive output and task automation. We draw analogies to economic production functions, extending them to explicitly include the effects of AI and cognitive tools.

\subsection{Baseline Cognitive Productivity Model}

Let $P_C$ denote cognitive productivity, which depends on three main factors:
\begin{itemize}
    \item $C_H$ -- Human cognitive capacity (memory, reasoning, skill)
    \item $K$ -- Accessible knowledge (written, digital, learned)
    \item $E$ -- Cognitive effort or time expended
\end{itemize}

A generic production function (Cobb-Douglas form) can be written as:
\begin{equation}
    P_C = A \cdot C_H^{\alpha} K^{\beta} E^{\gamma},
\end{equation}
where:
\begin{itemize}
    \item $A$ is a baseline productivity constant (reflecting institutional/technological efficiency)
    \item $\alpha, \beta, \gamma > 0$ are elasticity coefficients such that $\alpha + \beta + \gamma = 1$
\end{itemize}

\subsection{Impact of Cognitive Tools: Writing and Printing}

Historically, tools like writing and printing increased accessible knowledge $K$:
\begin{equation}
    K' = (1 + \delta_W)K,
\end{equation}
\begin{equation}
    K'' = (1 + \delta_P)K',
\end{equation}
where:
\begin{itemize}
    \item $\delta_W > 0$ is the fractional increase in knowledge enabled by writing,
    \item $\delta_P > 0$ is the additional increase due to printing.
\end{itemize}

The new productivity after both revolutions is:
\begin{equation}
    P_{C, \text{print}} = A \cdot C_H^{\alpha} \left[ (1+\delta_P)(1+\delta_W)K \right]^{\beta} E^{\gamma}
\end{equation}

\subsection{Formalizing AI-Driven Cognitive Augmentation}

AI affects cognitive production differently. It increases both the effective cognitive capacity and amplifies effort by automating tasks. Let:
\begin{itemize}
    \item $\eta_C \geq 1$ -- AI's multiplier on cognitive capacity ($C_H$)
    \item $\eta_E \geq 1$ -- AI's multiplier on effective cognitive effort (e.g., due to automation)
\end{itemize}

The AI-augmented productivity is:
\begin{equation}
    P_{C,\text{AI}} = A \cdot (\eta_C C_H)^{\alpha} \left[ (1+\delta_P)(1+\delta_W)K \right]^{\beta} (\eta_E E)^{\gamma}
\end{equation}

Expanding:
\begin{equation}
    P_{C,\text{AI}} = A \cdot \eta_C^{\alpha} \eta_E^{\gamma} C_H^{\alpha} \left[ (1+\delta_P)(1+\delta_W)K \right]^{\beta} E^{\gamma}
\end{equation}

\subsection{AI as a General-Purpose Cognitive Technology}

Suppose AI is applied across $n$ distinct cognitive tasks indexed by $i$:
\begin{equation}
    P_{C,\text{AI}}^{\text{Total}} = \sum_{i=1}^n A_i \cdot (\eta_{C,i} C_{H,i})^{\alpha_i} \left[ (1+\delta_P)(1+\delta_W)K_i \right]^{\beta_i} (\eta_{E,i} E_i)^{\gamma_i},
\end{equation}
where all parameters can be task-dependent.

\subsection{Marginal Productivity Analysis}

To measure AI's revolutionary potential, we analyze marginal productivities with respect to human cognition and effort:

\paragraph{Marginal Productivity with respect to Cognitive Capacity:}
\begin{equation}
    MP_{C_H} = \frac{\partial P_{C,\text{AI}}}{\partial C_H} = \alpha A \cdot \eta_C^{\alpha} \eta_E^{\gamma} C_H^{\alpha-1} \left[ (1+\delta_P)(1+\delta_W)K \right]^{\beta} E^{\gamma}
\end{equation}

\paragraph{Marginal Productivity with respect to Cognitive Effort:}
\begin{equation}
    MP_{E} = \frac{\partial P_{C,\text{AI}}}{\partial E} = \gamma A \cdot \eta_C^{\alpha} \eta_E^{\gamma} C_H^{\alpha} \left[ (1+\delta_P)(1+\delta_W)K \right]^{\beta} E^{\gamma-1}
\end{equation}

\subsection{Revolutionary Threshold Condition}

A productivity revolution is defined as a regime where AI-enhanced marginal productivities surpass a critical threshold $\tau$ compared to their pre-AI values:
\begin{equation}
    \frac{MP_{C_H}^{\text{AI}}}{MP_{C_H}^{\text{pre-AI}}} \geq \tau
\end{equation}
\begin{equation}
    \frac{MP_{E}^{\text{AI}}}{MP_{E}^{\text{pre-AI}}} \geq \tau
\end{equation}

For both, plugging in the ratios:
\begin{equation}
    \frac{MP_{C_H}^{\text{AI}}}{MP_{C_H}^{\text{pre-AI}}} = \eta_C^{\alpha} \eta_E^{\gamma}
\end{equation}
So the revolution threshold becomes:
\begin{equation}
    \eta_C^{\alpha} \eta_E^{\gamma} \geq \tau
\end{equation}

\subsection{AI-Driven Automation: Task Substitution Model}

Let $\mathcal{T}$ be the set of all cognitive tasks in a domain. For each task $t \in \mathcal{T}$, define a function $A(t)$ indicating if the task is automated by AI ($A(t)=1$) or performed by a human ($A(t)=0$). Then, total cognitive output is:
\begin{equation}
    P_{\text{total}} = \sum_{t \in \mathcal{T}} \left[ (1-A(t)) \cdot p_H(t) + A(t) \cdot p_{AI}(t) \right],
\end{equation}
where $p_H(t)$ is productivity for humans on task $t$, and $p_{AI}(t)$ is for AI.
If $p_{AI}(t) \gg p_H(t)$ for many $t$, overall productivity undergoes a step change—mathematically formalizing the “revolution.”

\subsection{Summary}

These equations capture the shift from traditional, human-constrained cognitive productivity to an AI-augmented regime. In this regime, both the scale and marginal gains from knowledge work are fundamentally changed, supporting the theoretical argument of a cognitive productivity revolution.

\section{Modern AI Applications Driving Cognitive Productivity}
AI's role as a cognitive engine is already evident in a wide array of modern applications. Here we highlight several domains where AI systems are measurably boosting productivity by performing or enhancing cognitive tasks that traditionally required human intelligence:

1. \textbf{Generative AI and Knowledge Work}: Large Language Models (LLMs) such as OpenAI's GPT-3 and GPT-4 have demonstrated an extraordinary ability to produce human-like text, write computer code, and even pass professional exams. These models serve as general-purpose cognitive assistants. For example, GPT-4 has passed legal bar exams and medical licensing exams at a level comparable to competent human test-takers, suggesting it can perform sophisticated reasoning and knowledge application in those fields. The immediate productivity implications are significant - routine drafting of emails, reports, or code can be offloaded to AI, allowing human workers to focus on higher-level tasks. The rapid adoption of such tools reflects their utility: ChatGPT (based on GPT-3.5 and GPT-4) reached 100 million users in about two months, an unprecedented rate of technology uptake
\cite{mehta2023chatgpt}
. This indicates that millions of people are already leveraging AI for tasks like writing, brainstorming, translating, and tutoring. In essence, generative AI is becoming a cognitive force multiplier for knowledge workers.

2. \textbf{AI in Healthcare Diagnostics}: AI systems are augmenting and, in some cases, rivaling human experts in medical diagnosis and analysis. A striking example comes from dermatology: a 2017 Nature study found that a deep convolutional neural network trained on 129,000 skin lesion images could diagnose skin cancer at dermatologist-level accuracy, even slightly outperforming a panel of 21 board-certified dermatologists in identifying malignant lesions in a controlled test
\cite{carlson2025ai}
. In radiology, AI models can analyze X-rays or MRI scans to detect abnormalities (like tumors or fractures) often as well as or faster than human radiologists. These tools can dramatically increase productivity by handling initial image screenings, flagging potential issues for human doctors, and reducing diagnostic errors. In practice, AI-aided diagnostics can mean earlier detection of diseases (improving outcomes) and time saved for healthcare professionals. Hospitals are beginning to deploy AI for tasks such as reading medical images, analyzing patient data to predict risks, and even automating routine documentation, thereby allowing doctors to spend more time on patient care - a cognitive reallocation of effort.

3. \textbf{Legal Automation and AI in Law}: The legal field, historically a labor-intensive knowledge profession, is being transformed by AI-based automation of research and document review. One notable benchmark involved an AI system tested against corporate lawyers in reviewing legal contracts (specifically Non-Disclosure Agreements for risks and issues). The result: the AI achieved 94\% accuracy, matching the best human lawyer, while the group of 20 experienced lawyers averaged 85\% accuracy
\cite{wood2018ai}
. Moreover, the AI took only about 26 seconds to complete what took human lawyers 92 minutes on average
\cite{wood2018ai}
. This stark speed advantage without loss of quality illustrates how AI can greatly boost productivity in legal work. AI-powered tools (such as legal research assistants and contract analysis platforms) can sift through vast databases of case law in seconds or automatically flag problematic clauses in contracts. Law firms and corporate legal departments are beginning to integrate such tools to handle due diligence, compliance checks, and other routine cognitive tasks. While complex legal reasoning and courtroom advocacy remain human domains for now, AI is reshaping the daily workflow of lawyers, enabling them to handle more cases or focus on strategy rather than rote review.

4. \textbf{AI in Scientific Discovery and Drug Development}: AI is accelerating discovery in science and medicine by functioning as a tireless researcher. In drug discovery, for instance, machine learning models can comb through enormous chemical space to identify promising new therapeutic molecules much faster than traditional trial-and-error lab work. A landmark example is the discovery of the antibiotic Halicin. In 2020, researchers at MIT employed a deep learning model to screen over 100 million chemical compounds for potential antibiotics. In a matter of days, the AI identified a novel molecule (later named Halicin) that proved effective against numerous highly resistant bacteria - including some ``untreatable'' strains - and cleared infections in mice
\cite{trafton2020ai}
. This was hailed as the first AI-discovered antibiotic
\cite{service2020ai,stokes2020deep}
. The AI essentially learned what molecular features make for a good antibiotic, then searched an immense space of possibilities to find candidates, a task infeasible for humans to do manually. Beyond chemistry, AI models (like DeepMind's AlphaFold) have solved grand challenges such as predicting protein structures from amino acid sequences - a breakthrough that can dramatically speed up biological research and drug design. In physics and engineering, AI systems are being used to suggest new materials or optimize complex designs (e.g. aerodynamic shapes, efficient batteries) by intelligently navigating design spaces. These examples show AI acting as a cognitive partner in innovation: generating hypotheses, running simulations, and interpreting data at a scale and speed well beyond unassisted human capability.

5. \textbf{Autonomous Agents and Systems}: AI's cognitive skills are also enabling autonomy in both digital and physical environments. Autonomous vehicles are a prime example - self-driving cars use AI perception and decision-making to chauffeur passengers without human input. Companies like Waymo have logged millions of miles with AI-driven cars, which must continuously analyze sensor data and make safe driving decisions, a cognitive task that requires visual understanding, prediction, and real-time planning. Similarly, autonomous drones and robots are being deployed in warehouses, hospitals, and factories, where they navigate and execute tasks by themselves. In the digital realm, AI agents can perform multi-step procedures such as booking appointments, responding to emails, or even conducting financial trades under certain rules. Early ``agentic'' AI systems (sometimes called AutoGPT and the like) demonstrate how an AI can be given a high-level goal and then break it down into sub-tasks, iteratively using its own outputs to achieve the goal - essentially showing a form of autonomous problem-solving. While still rudimentary, these agents hint at a future where repetitive cognitive workflows (filling forms, scheduling, basic data analysis) could be delegated entirely to an AI assistant. The productivity impact is akin to having a junior colleague or an autopilot for knowledge work that can handle background tasks. Taken together, from self-driving cars to virtual assistants, autonomous AI systems expand the scale of what can be handled without direct human oversight, again freeing humans to focus on supervision or more complex decisions.

These examples underscore that AI's role is not hypothetical or limited to laboratory demos; it is already manifesting across the economy. AI excels especially at tasks that are data-intensive, routine, or highly complex in a narrow domain. In many such tasks, AI brings not only speed, but often enhanced quality or new capabilities (e.g., finding patterns no human would notice). This directly translates into productivity growth: more output can be generated per unit of human time, or entirely new outputs become feasible. However, these gains also come with challenges and choices, particularly around how AI is integrated with human labor. The following sections will explore the broader economic and social implications of this cognitive revolution, asking: Will AI drive productivity in a way that broadly benefits society, as past revolutions eventually did? How do we ensure that amplifying cognitive work through AI leads to augmentation of human potential rather than mere replacement?

\section{Economic Perspectives: AI, Productivity and Labor}
From an economic standpoint, the rise of AI as a cognitive engine raises both optimism and caution. Historically, major technological revolutions (steam power, electricity, computing) have eventually led to higher productivity and prosperity, but often after a period of adjustment and sometimes with uneven effects on labor. AI, being a general-purpose cognitive technology, is expected to significantly boost productivity - but the magnitude and distribution of its benefits depend on how we direct and implement it.

1. \textbf{AI as a General-Purpose Technology}: Economists Erik Brynjolfsson and Andrew McAfee have characterized AI, particularly modern machine learning, as a general-purpose technology similar to past engines of growth (like the internal combustion engine or electricity). Such technologies have wide-ranging uses, improve over time, and spur complementary innovations. A report by Goldman Sachs, for example, projected that AI adoption could raise global productivity growth by 1.5 percentage points annually and increase global GDP by around 7\% (some \$7 trillion in output) over a decade
\cite{acemoglu2023rebalancing}
. These rosy forecasts assume that AI will be deployed widely to create efficiency gains across sectors. Indeed, if AI can handle cognitive tasks, it can potentially make every knowledge worker - from customer service agents to researchers - more productive, just as mechanization made manufacturing workers more productive in the past. However, realizing these gains isn't automatic.

2. \textbf{The Productivity Paradox}: There is a well-known phenomenon where new technologies initially do not show up in productivity statistics, often called the Solow paradox or productivity paradox. We saw this with computers in the 1970s-80s: despite rapid advances in computing, productivity growth slowed, leading economist Robert Solow to quip, ``You can see the computer age everywhere but in the productivity statistics.'' A similar discussion is happening around AI. Some recent data has shown tepid productivity growth in many economies even as AI algorithms improve. One reason, as economic historians note, is that it takes time to reorganize workflows and institutions to fully harness a new technology's potential. For example, when electricity was introduced into factories in the late 19th century, initial gains were small because factories simply swapped steam engines for electric motors without changing the floor layouts designed for steam power. Only after factories redesigned their entire workflows (for instance, moving from vertical multi-floor layouts to single-floor assembly lines powered by distributed electric motors) did productivity soar
\cite{tinholt2025productivity}
. By analogy, firms today might install AI systems to automate existing processes, but if they don't fundamentally rethink business models and workflows to leverage AI's unique strengths (e.g. its ability to analyze vast data or make quick predictions), the productivity gains may remain limited. Economic analysts argue that the real leap will come from reimagining operations around AI rather than just ``pasting'' AI onto old processes
\cite{tinholt2025productivity}
. This might entail changes like redesigning supply chains with AI-driven predictive ordering, or reorganizing teams to integrate AI decision-support in every project.

3. \textbf{Automation vs Augmentation - The Labor Question}: A central economic debate is whether AI will predominantly replace human labor (automation) or augment it, and what that means for employment and wages. On one hand, if AI takes over many tasks, it could displace workers in those functions (similar to how industrial machines displaced some artisans). On the other hand, if AI complements human workers - taking over the tedious tasks and enabling humans to focus on more complex ones - it could enhance workers' productivity and potentially create new jobs (just as the ATM automated bank tellers' routine tasks but led them to focus on customer relationships, resulting in more bank branches and employment overall in banking). Which of these scenarios dominates is not predetermined; it depends on technological design choices, business strategies, and policies
\cite{acemoglu2023rebalancing}
.

Some economists warn that the current trajectory of AI is too tilted towards automation of human roles without creating enough new roles. Daron Acemoglu and Simon Johnson, in their book Power and Progress, note that many AI applications so far have been ``so-so automation'' - they replace humans in certain tasks but don't dramatically improve productivity enough to generate large new wealth or jobs. They argue that simply letting AI develop along a path of maximal automation (what they dub a ``just let AI happen'' approach) is risky and not guaranteed to produce broad economic gains
\cite{acemoglu2023rebalancing}
. Indeed, Acemoglu's research shows that while automation does typically increase output per worker (average productivity), it may not increase marginal productivity of labor, meaning firms have less incentive to hire or pay workers more
. Over the last few decades, automation technologies (like industrial robots or software) have contributed to rising productivity and corporate profits, but wage growth has stagnated for many workers and labor's share of income has fallen
\cite{acemoglu2023rebalancing}
. In other words, the gains often accrue to capital owners or a few high-skilled workers, exacerbating inequality.

Brynjolfsson offers a complementary perspective with the concept of the ``Turing Trap'': he cautions that excessive focus on making AI imitate and replace humans (achieving human-like AI in a narrow sense) could lead to a trap where humans lose bargaining power and income, as machines become substitutes for their labor
\cite{brynjolfsson2022turing}
. By contrast, if AI is developed and deployed to augment human abilities - working with people rather than instead of them - it can empower workers and lead to more widely shared prosperity
. Augmenting AI would mean creating systems that partner with humans, enabling new tasks and higher productivity with human input. An example is using AI as a diagnostic assistant for doctors: the AI doesn't replace the doctor but makes them more effective, potentially allowing them to see more patients or achieve better outcomes. This augmentation approach not only preserves human roles but can even create new ones (for instance, demand for AI-savvy medical practitioners, or entirely new job categories we can't yet imagine). Importantly, Brynjolfsson argues that augmentation tends to generate more total value than pure automation
, because it opens up new services and improvements rather than just cutting costs. However, he observes that ``there are currently excess incentives for automation rather than augmentation'' in how businesses and research are oriented
\cite{brynjolfsson2022turing}
 - it is often easier and more immediately profitable to design AI to cut labor costs than to create new tasks for humans.

The policy implication, as Acemoglu, Johnson, and Brynjolfsson all suggest, is that society can and should shape the AI revolution's path. There is nothing inevitable about AI purely eroding jobs or, conversely, about it ushering in shared prosperity - it depends on choices. For instance, incentives (like tax breaks or R\&D funding) could encourage AI that complements workers (think AI tools for teachers to personalize education, rather than AI that replaces teachers). Training programs and education can prepare workers to use AI tools effectively (more on that in the next section). If left purely to market forces with a narrow cost-cutting lens, we might get a wave of automation that boosts productivity but concentrates its gains. But with conscious effort, AI's cognitive capabilities could be harnessed in ways that broadly raise human productivity and create new forms of employment. History shows both sides: the Industrial Revolution initially caused displacement and misery for many (e.g., weavers replaced by mechanized looms), but eventually new industries, jobs, and higher living standards emerged. The difference now is forethought - economists are actively discussing how to avoid a long period of inequality or jobless growth by guiding AI development. 
 
Another economic dimension is how AI might change the nature of work itself. As routine cognitive tasks are automated, the comparative advantage of human labor may shift to areas that require creativity, complex problem-solving, interpersonal skills, and other forms of ``soft” cognition that AI is not yet good at (and perhaps never will be, if they involve emotional intelligence, ethical judgment, or genuinely novel thinking). This echoes what happened with physical automation: machines took over repetitive factory work, and human labor shifted towards service, creativity, and coordination tasks. We might see a similar shift within cognitive work: mundane data processing or straightforward analysis is handled by AI, while humans focus on intuitive decision-making, cross-domain thinking, or tasks requiring human contact. This could increase the economic value of certain skills and roles (like managers who can integrate AI outputs into strategy, or designers who use AI tools to enhance creativity) while diminishing demand for some entry-level analytical jobs that could be fully automated. 
 
In summary, the economic outlook on AI-driven productivity is one of great potential tempered by the need for adaptation. AI undeniably has the technical capacity to raise output and efficiency in many processes - often dramatically so - as evidenced by the examples earlier. The open question is how those efficiency gains translate into economic welfare: Will they mostly cut costs for firms and increase profits? Will they lead to cheaper goods and services for consumers (a plus for living standards)? Will they free up human workers to do more meaningful and higher-paying work, or will many find themselves displaced? The answers will depend on how we navigate the transition. The next section turns to the social and institutional response required to maximize the positive impact of this cognitive revolution.

\section{Sociological Perspectives: Knowledge Labor and Institutional Adaptation}
As AI transforms the nature of work through cognitive automation, it also prompts profound changes in the social fabric of workplaces, educational systems, and institutions at large. Understanding these changes requires a sociological lens: How do we adapt our roles, skills, and organizations to integrate AI in a human-centric way? What happens to the ``knowledge workers'' of the 20th century in an era where machines can perform knowledge work?

1. \textbf{Transformation of Knowledge Labor}: The late 20th century saw the rise of the ``knowledge economy,'' where a large portion of the workforce engaged in jobs centered on information processing, analysis, and creative thinking rather than manual labor. AI's encroachment into this realm means that many such knowledge jobs will be redefined. Rather than rendering human knowledge workers obsolete, evidence so far suggests a reconfiguration of tasks. For instance, journalists now often use AI to generate basic news reports or summarize data, freeing them to focus on investigative pieces or complex storytelling. Financial analysts might rely on AI to crunch numbers and detect patterns, while they concentrate on strategy and client communication. In software development, tools like AI code assistants (e.g. GitHub Copilot) can automatically write routine code or detect errors, allowing developers to spend more time on design and creative problem-solving. This dynamic - AI handling the boilerplate, humans handling the novel and the interpersonal - is likely to become the norm in many professions.

2. \textbf{Institutional Adaptation - The Workplace}: For organizations, adopting AI is not just a technical upgrade but an organizational change. Companies that successfully leverage AI often must redesign workflows, retrain employees, and even change their structure. A common recommendation is that companies identify which tasks in each role can be automated and which should be enhanced. They may then redefine job descriptions so that employees explicitly work alongside AI tools. For example, a consulting firm might require its analysts to use an AI research assistant for gathering information, and spend more time on synthesizing insights for clients. This could change hiring criteria (more emphasis on creative thinking and less on rote analytical ability, since the latter is handled by AI). Some firms might adopt a ``center of excellence'' model for AI, creating dedicated teams that integrate AI solutions into various departments, ensuring consistency and addressing ethical or quality concerns. We also see institutions developing AI governance frameworks - internal policies on how and when to use AI in decision-making, to maintain accountability and trust.

3. \textbf{Institutional Adaptation - Policy and Society}: On a broader societal level, institutions such as governments and regulatory bodies face the task of adapting to AI's spread. This includes updating laws (for example, how to handle liability when an AI-driven autonomous vehicle causes an accident, or how to regulate AI in sensitive areas like finance or healthcare). There is also an increasing recognition of the need for social safety nets and continuous learning systems to support workers who may be displaced or need to transition roles. Discussions about universal basic income or wage insurance have been partly motivated by the potential of AI-induced job disruptions. Additionally, labor institutions (like unions and professional associations) are negotiating how AI should be deployed - for instance, calling for ``human-in-the-loop'' guarantees or for shared productivity gains. An example is the writers' and actors' unions in Hollywood negotiating over the use of AI in script writing or digital actor replicas, which shows the preemptive steps being taken to ensure AI doesn't undermine creative labor's value.

4. \textbf{Cultural Adaptation}: Human attitudes toward AI at work range from excitement to anxiety. Some workers fear that reliance on AI could deskill their professions or even make their roles redundant. There is also a concern about losing the human touch or craftsmanship in various fields (for instance, if AI writes code, will programmers lose the deeper understanding of their craft over time?). These fears echo Socrates' worry about writing, mentioned earlier, and similar fears in every technological shift (e.g., calculators ruining arithmetic skills). Research on ``skill decay'' suggests that if people stop practicing certain cognitive skills because AI handles them, those skills do atrophy
\cite{anseel2025ai}
. For example, if one relies on AI for writing, one might lose some ability to compose thoughts clearly unaided. Sociologically, this means we might need to deliberately retain and cultivate foundational skills even as AI provides shortcuts - much like education still teaches arithmetic despite calculators, because the conceptual understanding is valuable. Some have advocated for a kind of ``bilingual'' ability in both human and AI problem-solving: humans should understand how to solve problems manually to some extent, even if they often delegate to AI, to avoid total dependency and to better oversee the AI's output.

In response to these issues, there are emerging norms and strategies: companies pairing new AI deployments with training sessions, society emphasizing media literacy and critical thinking (so people can judge AI-generated content), and ethical guidelines (e.g., requiring AI to explain its reasoning in high-stakes decisions to maintain transparency). Institutions of science and knowledge production are also adapting - for example, academic journals now grapple with policies on AI-generated text or images in submissions, trying to maintain integrity and credit human intellectual contribution appropriately.

In summary, the sociological dimension of the AI-driven cognitive revolution involves a broad spectrum of adaptations: workers adapting their skill sets, organizations adapting their structures and cultures, and society adapting its policies and expectations. The guiding principle widely suggested is to focus on human-AI collaboration. As the American Academy of Dermatology's Augmented Intelligence Committee chair Ivy Lee remarked, ``we are now past the hype… we're questioning how we can use [AI] practically and meaningfully in the real world''
\cite{carlson2025ai}
 - the emphasis is on finding the role for AI that best complements human strengths. Institutions that successfully integrate AI are likely those that treat it not as a human replacement, but as a powerful new partner and tool that, when wielded correctly, can elevate human productivity to new heights.

\section{Conclusion}
\label{sec:conclusion}
The rise of AI as a cognitive engine marks a pivotal juncture in the story of human productivity. We have argued that AI should be viewed not as just another automation tool, but as the harbinger of a new kind of productivity revolution - one grounded in the mechanics of thought and knowledge rather than the mechanics of industry. This revolution carries the transformative force of the great leaps of the past (agriculture, writing, printing, industrialization), yet it is unfolding on a faster, broader scale. 

\textbf{By comparing AI's emergence to the advent of written language and the printing press, we highlighted the continuity in humanity's quest to expand our cognitive reach.} Writing allowed us to store and share knowledge externally; printing allowed that knowledge to propagate widely; now AI allows knowledge to be processed, synthesized, and even created with machine assistance. Each step magnified what individuals and societies could accomplish intellectually. AI, in effect, is extending the frontier of human cognition, enabling tasks to be done in seconds by algorithms that would take humans years (or be outright impossible for us). From composing coherent text, to diagnosing complex diseases, to discovering new scientific insights, AI is proving itself as an engine that turns data into actionable intelligence. 

The empirical examples reviewed - spanning GPT's conversational genius to AI's feats in medicine, law, and science - illustrate that this is not a theoretical future but a present reality. Even so, we are likely in the early days of the cognitive revolution. Just as the first steam engines in the 18th century were primitive compared to what the Industrial Revolution eventually achieved, today's AI, impressive as it is, may be a precursor to far more potent cognitive machines in coming decades. This prospect makes it all the more crucial to learn from history and proactively shape outcomes. 

The multidisciplinary perspectives considered in this paper converge on a common theme: the importance of human-centered integration of AI. \textbf{Economic analyses remind us that productivity gains alone do not guarantee widespread prosperity - it matters how those gains are obtained and distributed.} If we choose an augmentation path, where AI amplifies human capabilities, we increase the likelihood of broad-based benefits: higher productivity accompanied by new job opportunities, higher wages, and enriched human work. If instead we drift into a pure automation path, we risk a scenario of concentrated gains and disempowered labor. The encouraging news is that AI doesn't inherently prefer one path or the other; it is a tool we design and deploy. As Brynjolfsson observed, when AI is used to augment rather than imitate humans, it can generate new capabilities, new products, and more total value
\cite{brynjolfsson2022turing}
. This suggests a strategic vision: focus AI on solving problems that expand what humans can do (curing diseases, educating more people, addressing climate challenges) rather than just cutting costs. 

From a social and institutional angle, the onus is on us to adapt and evolve alongside our new cognitive tools. \textbf{The cognitive revolution need not be a zero-sum game of humans versus machines. Instead, it can be a symbiosis - a phase where we reorganize our institutions (workplaces, schools, laws) to leverage the best of both human and artificial intelligence.} We must invest in human capital - training people in the skills that complement AI, and cultivating the uniquely human traits of creativity, empathy, and critical thinking. As one healthcare education article noted, this era makes education ``not only about knowledge transfer but also about enhancing human cognitive capabilities to navigate and shape a rapidly evolving technological landscape”
\cite{zhang2024preparing}
. Lifelong learning and adaptability will become central virtues of the workforce. Crucially, we should maintain a clear-eyed optimism balanced with vigilance. AI's potential for good - from saving lives via better medical diagnoses to democratizing knowledge via personal tutors - is immense. But to realize that potential, broad societal engagement is needed. Ethicists, sociologists, economists, and technologists must continue to collaborate to ensure AI is developed responsibly and inclusively. Issues like algorithmic bias, privacy, and security of AI systems must be addressed to build trust. At the same time, access to AI tools should be expanded so that their benefits are widely shared (for instance, not only big corporations but also small businesses, schools, and developing countries harnessing AI for growth). Institutional innovations, such as new governance models for AI or international agreements on its use, may be necessary to steer this revolution in beneficial directions. 

In conclusion, AI as a cognitive engine offers a profound opportunity to reinvent productivity in a way that plays to the highest human strengths. It is as if we have created a “steam engine for the mind” - a machine that can carry part of the cognitive load. Just as steam power freed humans from some drudgery of manual labor, cognitive AI has the potential to free us from some mental drudgery, allowing humanity to focus more on what matters: creativity, discovery, and connection. The difference is that this time, the “machine” is our partner in thinking. 

Whether this cognitive revolution ultimately mirrors the success of the industrial revolution - in raising living standards and opening new frontiers - will depend on choices we make now. By anchoring AI's development in a human-centered framework, by redesigning our institutions to adapt to and capitalize on AI, and by proactively managing the transition for workers and communities, we can ensure that AI as language, not steam truly becomes a tool for human advancement. In doing so, we might look back and recognize this era as the moment when humanity didn't just build smarter machines, but when we learned to work smarter with machines - unleashing a new wave of innovation and productivity that carries all of society forward.

\bibliographystyle{plain}
\bibliography{sample-base}

\end{document}